\documentclass[journal]{IEEEtran}

\usepackage[english]{babel}

\usepackage{multirow}
\usepackage[numbers,sort&compress,square,comma]{natbib}

\ifCLASSINFOpdf
  \usepackage[pdftex]{graphicx}
  \DeclareGraphicsExtensions{.pdf,.jpeg,.png}
\else
  \usepackage[dvips]{graphicx}
  \DeclareGraphicsExtensions{.eps}
\fi
\usepackage{amsmath}

\begin{document}
\title{Metric Learning in Codebook Generation of Bag-of-Words for Person Re-identification}
%
%
%

\author{Lu~Tian,~\IEEEmembership{Student~Member,~IEEE,}
        and~Shengjin~Wang,~\IEEEmembership{Member,~IEEE}%
\thanks{Lu Tian and Shengjin Wang are with the Department of Electronic Engineering, Tsinghua University, Beijing, 100084 China e-mail: tl.september@gmail.com.}%
\thanks{Manuscript received April 19, 2005; revised August 26, 2015.}}

%
%

\markboth{Journal of \LaTeX\ Class Files,~Vol.~14, No.~8, August~2015}%
{Lu Tian \MakeLowercase{\textit{et al.}}: Metric Learning in Codebook Generation of Bag-of-Words for Person Re-identification}
%



\maketitle

\begin{abstract}
Person re-identification is generally divided into two part: first how to represent a pedestrian by discriminative visual descriptors and second how to compare them by suitable distance metrics. Conventional methods isolate these two parts, the first part usually unsupervised and the second part supervised. The Bag-of-Words (BoW) model is a widely used image representing descriptor in part one. Its codebook is simply generated by clustering visual features in Euclidian space. In this paper, we propose to use part two metric learning techniques in the codebook generation phase of BoW. In particular, the proposed codebook is clustered under Mahalanobis distance which is learned supervised. Extensive experiments prove that our proposed method is effective. With several low level features extracted on superpixel and fused together, our method outperforms state-of-the-art on person re-identification benchmarks including VIPeR, PRID450S, and Market1501.
\end{abstract}


%
\IEEEpeerreviewmaketitle

\section{Introduction}
%
%
%
%
\IEEEPARstart{P}{erson} re-identification \cite{gong2014person} is an important task in video surveillance systems. The key challenge is the large intra-class appearance variations, usually caused by various human body poses, illuminations, and different camera views. Furthermore, the poor quality of video sequences makes it difficult to develop robust and efficient features.

Generally speaking, person re-identification can be divided into two parts: first how to represent a pedestrian by discriminative visual descriptors and second how to compare them by suitable distance metrics. Bag of words (BoW) model and its variants is one of the most widely used part one image descriptor technology in person re-id systems with significant performance \cite{lu2015person}. In the traditional BoW approaches, images are divided into patches and local features are first extracted to represent these patches. Then a codebook of visual words is generated by unsupervised clustering. After that, the image is represented by histogram vectors obtained by mapping and quantizing the local features into the visual words in the codebook.

However, it is not optimal to cluster visual words by k-means in Euclidian space, which implicitly assumes that local features of the same person usually have closer Euclidian distance, which does not always stand in practical.

Part two metric learning methods learn suitable distance metrics of image descriptors to distinguish correct and wrong matching pairs. However, conventional methods always isolate part one and part two, the first part usually unsupervised and the second part supervised.

\begin{figure}[!t]
\centering
\includegraphics[width=2.5in]{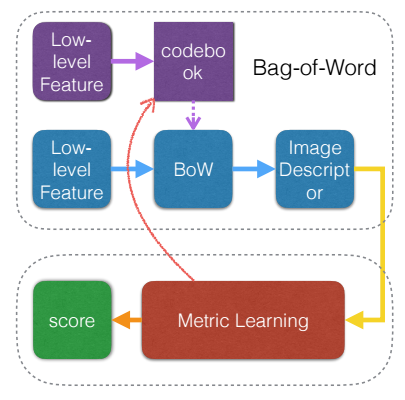}
\caption{The framework of metric learning in codebook generation of Bag-of-Words.}
\label{fig_framework}
\end{figure}

To this end, this paper proposes to borrow some part two metric learning techniques to learn a suitable distance for local features in part one BoW model. In particular, a Mahalanobis distance is trained on local features extracted from pedestrian images. Then codebook of visual words is clustered under this Mahalanobis distance. We formulate the codebook generation task as a distance metric learning problem and propose to use KISSME \cite{kostinger2012large} to solve it efficiently. When integrated with conventional part two metric learning methods, our proposed method also achieves good performance. The overall framework of our proposed method is shown in Fig \ref{fig_framework}. Finally, we outperform state-of-the-art result by applying KISSME \cite{kostinger2012large} metric learning for local features in the BoW model and Null Space \cite{zhang2016learning} metric learning for image descriptors after the BoW model.

In summary, our contributions are three-fold: 1), to the best of our knowledge, we are the first to propose metric learning for BoW low level features; 2), we propose KISSME \cite{kostinger2012large} to learn a suitable metric for low level features; 3) we integrate the proposed local feature level metric learning method with conventional part two image descriptor level metric learning methods and achieve state-of-the-art results.

The rest of this paper is organized as follows. In Section \ref{s:related_work}, a brief discussion of several related works on person re-identification is made. In Section \ref{s:approach} we introduce our method. The experimental results are shown and examined in Section \ref{s:experiments}. Finally, we draw our conclusions in Section \ref{s:conclusion}.

\section{Related Work}
\label{s:related_work}

Generally speaking, person re-id include two basic parts: how to represent a pedestrian and how to compare them, and most efforts on person re-id could be roughly divided into these two categories~\cite{zheng2016person}.

The first category focuses on discriminative visual descriptor extraction. \citeauthor{gray2008viewpoint} \cite{gray2008viewpoint} use RGB, HS, and YCbCr color channels and 21 texture filters on luminance V channel, and partition pedestrian images into horizontal strips. \citeauthor{farenzena2010person} \cite{farenzena2010person} compute a symmetrical axis for each body part to handle viewpoint variations, based on which the weighted color histogram, the maximally stable color regions, and the recurrent high-structured patches are calculated. \citeauthor{zhao2013unsupervised} \cite{zhao2013unsupervised} propose to extract 32-dim LAB color histogram and 128-dim SIFT descriptor from each 10*10 patch. \citeauthor{das2014consistent} \cite{das2014consistent} use HSV histograms on the head, torso and legs. \citeauthor{li2013learning} \cite{li2013learning} aggregate local color features by hierarchical Gaussianization \cite{zhou2009hierarchical,chen2015similarity} to capture spatial information. \citeauthor{pedagadi2013local} \cite{pedagadi2013local} extract color histograms from HSV and YUV spaces and then apply PCA dimension reduction. \citeauthor{liu2014semi} \cite{liu2014semi} extract HSV histogram, gradient histogram, and the LBP histogram from each patch. \citeauthor{yang2014salient} \cite{yang2014salient} propose the salient color names based color descriptor (SCNCD) and different color spaces are analyzed. In \cite{liao2015person}, LOMO is proposed to maximize the occurrence of each local pattern among all horizontal sub-windows to tackle viewpoint changes and the Retinex transform and a scale invariant texture operator are applied to handle illumination variations. In \cite{lu2015person}, Bag-of-Words (BoW) model is proposed to aggregate the 11-dim color names feature \cite{van2007learning} from each local patch.

The second category learns suitable distance metrics to distinguish correct and wrong match pairs. Specifically, most metric learning methods focus on Mahalanobis form metrics, which generalizes Euclidean distance using linear scaling and rotation of the feature space, and the distance between two feature vectors \(x_{i}\) and \(x_{j}\) could be written as \[s(x_{i}, x_{j})=\sqrt{(x_{i}-x_{j})^{T} \textbf{M} (x_{i}-x_{j})},\] where \(\textbf{M}\) is the positive semi-definite Mahalanobis matrix. \citeauthor{weinberger2009distance} \cite{weinberger2009distance} propose the large margin nearest neighbor learning (LMNN) which sets up a perimeter for correct match pairs and punishes those wrong match pairs. In \cite{kostinger2012large}, KIEEME is proposed under the assumption that \(x_{i}-x_{j}\) is a Gaussian distribution with zero mean. \citeauthor{hirzer2012relaxed} \cite{hirzer2012relaxed} obtained a simplified formulation and a promising performance by relaxing the positivity constraint required in Mahalanobis metric learning. \citeauthor{li2013learning} \cite{li2013learning} propose locally-adaptive decision functions (LADF) combining a global distance metric and a locally adapted threshold rule in person verification. \citeauthor{chen2015similarity} \cite{chen2015similarity} add a bilinear similarity in addition to the Mahalanobis distance to model cross-patch similarities. \citeauthor{liao2015efficient} \cite{liao2015efficient} propose weighting the positive and negative samples differently. In \cite{liao2015person}, XQDA is proposed as an extension of Bayesian face and KISSME, in that a discriminant subspace is further learned together with a distance metric. It learns a projection \(w\) to the low-dimensional subspace in a similar way as linear discriminant analysis (LDA) \cite{scholkopft1999fisher} with \[\mathcal{J}(w) = \frac{w^{T}S_{b}w}{w^{T}S_{w}w}\] maximized, where \(S_{b}\) is the between-class scatter matrix and \(S_{w}\) is the within-class scatter matrix. \citeauthor{zhang2016learning} \cite{zhang2016learning} propose Null Space to further employ the null Foley-Sammon transform to learn a discriminative null space with the projection \(w\) where the within-class scatter is zero and between-class scatter is positive, thus maximizing \(\mathcal{J}(w)\) to positive infinite.

Recently some works based on deep learning are also used to tackle person re-id problem. Filter pairing neural network (FPNN) ~\cite{li2014deepreid} is proposed to jointly handle misalignment, photometric and geometric transforms, occlusions and background clutter with the ability of automatically learning features optimal for the re-identification task. Ahmed et al. ~\cite{ahmed2015improved} present a deep convolutional architecture and propose a method for simultaneously learning features and a corresponding similarity metric for person re-identification. Compared to hand-crafted features and metric learning methods, Yi et al. ~\cite{yi2014deep} proposes a more general way that can learn a similarity metric from image pixels directly by using a "siamese" deep neural network. A scalable distance driven feature learning framework based on the deep neural network is presented in ~\cite{ding2015deep}. ~\citeauthor{zheng2016discriminatively} ~\cite{zheng2016discriminatively} propose a new siamese network that simultaneously computes identification loss and verification loss, which learns a discriminative embedding and a similarity measurement at the same time. Pose invariant embedding (PIE) is proposed as a pedestrian descriptor in ~\cite{zheng2017pose}, which aims at aligning pedestrians to a standard pose to help re-id accuracy.

\section{The Approach}
\label{s:approach}

\subsection{Review of Bog-of-Words in Person Re-identification}

The BoW model represents an image as a collection of visual words. We briefly review the BoW model in person re-identification in previous approaches \cite{lu2015person,zheng2015scalable}. First, an pedestrian image \(i\) is segmented as superpixels by SLIC method \cite{achanta2012slic}. Superpixel algorithms cluster pixels into perceptually meaningful atomic regions according to the pixel similarity of color and texture, which capture image redundancy and provide a convenient primitive to compute robust image features. To enhance geometric constraints, the pedestrian image is usually partitioned into horizontal strips with equal width. Then in superpixel \(k\) of strip \(j\), the low level high-dimensional appearance features are extracted as \(\textbf{f}_{i,j,k} \in \mathcal{R}^{d}\) and \(d\) is the feature vector length. These low level features may contain much noise and redundancy, and are often difficult to use directly. Hence, a codebook \(\mathcal{C}=\{\textbf{c}(l)\}\) of visual words is generated by clustering (usually standard k-means) on these features and each word \(\textbf{c}\) corresponds to a cluster center with \(l\) in a finite index set. The mapping, termed as a quantizer, is denoted by: \(\textbf{f} \rightarrow \textbf{c}(l(\textbf{f}))\). The function \(l(\cdot)\) is called an encoder, and function \(\textbf{c}(\cdot)\) is called a decoder \cite{gray1984vector}. The encoder \(l(\textbf{f})\) maps any \(\textbf{f}\) to the index of its nearest codeword in the codebook \(\mathcal{C}\). Here multiple assignment (MA) \cite{jegou2008hamming} is employed, where the local feature \(\textbf{f}_{i,j,k}\) is assigned to some of the most similar visual words by measuring the distance between them. Thus the histogram of the visual words representing strip \(j\) is obtained by encoding the local features into the codebook, which is denoted as \(\textbf{d}_{i,j}=histogram\{l(\textbf{f}_{i,j,k})|k \in {strip}_{j} \}\). Each visual word is generally weighted using the TF scheme [2], [3]. We also use pedestrian parsing and background extraction techniques \cite{luo2013pedestrian} and only the superpixels which contain pedestrian parts are considered and counted in our BoW model. The BoW descriptor of image \(i\) is the concatenation of \(\textbf{d}_{i}=[\textbf{d}_{i,1},\cdots,\textbf{d}_{i,j},\cdots,\textbf{d}_{i,J}]\). Finally, the distance of two images \(i1\) and \(i2\) can be directly calculated as the Euclidean distance between \(\textbf{d}_{i1}\) and \(\textbf{d}_{i2}\), that is, \[s(i1, i2)=\sqrt{(\textbf{d}_{i1}-\textbf{d}_{i2})^{T} \cdot (\textbf{d}_{i1}-\textbf{d}_{i2})}.\] Or conventional part two metric learning methods can be applied to improve re-id performance by supervised labels. Most of them focus on Mahalanobis based metrics, which generalizes Euclidean distance using linear scalings and rotations of the feature space and can be written as \[s(i1, i2)=\sqrt{(\textbf{d}_{i1}-\textbf{d}_{i2})^{T} \textbf{M} (\textbf{d}_{i1}-\textbf{d}_{i2})},\] where \(\textbf{M}\) is the positive semi-definite Mahalanobis matrix.

Fusing different low level features together could provide more rich information. We consider four different appearance based features: color histograms (CH or namely HSV) \cite{lu2015person}, color names (CN) \cite{berlin1991basic,van2007learning}, HOG \cite{dalal2005histograms}, and SILTP \cite{liao2010modeling} to cover both color and texture characteristics. They are all \(l_{1}\) normalized followed by \(\sqrt{(\cdot)}\) operator before BoW quantization, as the Euclidean distance on root feature space is equivalent to the Hellinger distance on original feature space, and Hellinger kernel performs better considering histogram similarity \cite{arandjelovic2012three}. The fusion is applied at image descriptor level, which has been demonstrated effective. Different codebooks \(\mathcal{C}^{HSV}\), \(\mathcal{C}^{CN}\), \(\mathcal{C}^{HOG}\), and \(\mathcal{C}^{SILTP}\) are generated for each low level feature separately, thus the BoW image descriptor of each feature is calculated respectively. Then the final descriptor 
of image \(i\) is concatenated as \(\textbf{d}_{i}=[\textbf{d}_{i}^{HSV}, \textbf{d}_{i}^{CN}, \textbf{d}_{i}^{HOG}, \textbf{d}_{i}^{SILTP}]\).

\subsubsection{Color Histograms}
HSV is typically used to describe color characteristics within one region. First, the image is transferred to the HSV color space. Then the statistical distribution of hue (H) and saturation (S) channels is calculated respectively with each channel quantized to 10 bins. Luminance (V) channel is excluded because of huge illumination changes in person re-identification tasks.

\subsubsection{Color Names}
CN are semantic attributes obtained through assigning linguistic color labels to image pixels. Here, we use the descriptors learned from real-world images like Google Images to map RGB values of a pixel to 11 color terms \cite{van2007learning}. The CN descriptor assigns each pixel an 11-D vector, each dimension corresponding to one of the 11 basic colors. Afterward, the CN descriptor of a superpixel region is computed as the average value of each pixel.

\subsubsection{HOG}
HOG is a classical texture descriptor which counts occurrences of gradient orientation in localized portions of an image. We separate gradient orientation into 9 bins and calculate on the gray image.

\subsubsection{Scale Invariant Local Ternary Pattern}
SILTP \cite{liao2010modeling} descriptor is an improved operator over the well-known Local Binary Pattern (LBP) \cite{ojala1996comparative}. LBP has a nice invariant property under monotonic gray-scale transforms, however, it is not robust to image noises. SILTP improves LBP by introducing a scale invariant local comparison tolerance, achieving invariance to intensity scale changes and robustness to image noises. Within each superpixel, we extract 2 scales of SILTP histograms (\(SILTP^{0.3}_{4,3}\) and \(SILTP^{0.3}_{4,5}\)) as suggested in \cite{liao2015person}.

\subsection{Bag-of-Words Framework and Codebook Generation}

Codebook generation is a critical step of building the BoW model. Conventional approach simply clusters low level appearance features by unsupervised k-means in Euclidean space. In this paper, we suggest applying supervised metric learning methods and cluster features in Mahalanobis space with its trained distance metrics.

We denote the feature vector of superpixel \(k\) in the strip \(j\) of image \(i\) as \(\textbf{f}_{i,j,k}\), whereas \(\textbf{f}_{i,j,k} \in \mathcal{R}^{d}\) and \(d\) is the feature vector length. And (\(\textbf{f}_{i1,j,k1}\), \(\textbf{f}_{i2,j,k2}\)) is a pairwise feature instance where they belong to two superpixels in the same horizontal strip \(j\) of two different images. Here, only features belonging to the same horizontal strip are collected as pairwise instance, which is quite reasonable because of the geometric constrains of pedestrian images and dramatically reduce the amount of pairwise feature instances as well as the computational complexity. We further denote \(\mathcal{P}\) as the positive set of pairwise feature instances where the first feature and the second feature belong to same person, i.e, \((\textbf{f}_{i1,j,k1}, \textbf{f}_{i2,j,k2}) \in \mathcal{P}, id(i1)=id(i2)\). And we denote \(\mathcal{N}\) as the negative set of pairwise feature instances, i.e, \((\textbf{f}_{i1,j,k1}, \textbf{f}_{i2,j,k2}) \in \mathcal{N}, id(i1) \neq id(i2)\). The goal of our task is to learn a distance metric \(\textbf{M}'\) (to be distinguished with \(\textbf{M}\) in conventional part two metric learning methods) to effectively measure distance between any two visual features \(\textbf{f}_{i1,j,k1}\) and \(\textbf{f}_{i2,j,k2}\), which is often represented as \[d(\textbf{f}_{i1,j,k1}, \textbf{f}_{i2,j,k2})=\sqrt{(\textbf{f}_{i1,j,k1}- \textbf{f}_{i2,j,k2})^{T} \textbf{M}' (\textbf{f}_{i1,j,k1}- \textbf{f}_{i2,j,k2})},\] where matrix \(\textbf{M}'\) is the \(d \times d\) Mahalanobis matrix that must be positive and semi-definite.

Many metric learning methods are proposed to learn an optimized \(\textbf{M}'\). In this paper, we use KISSME \cite{kostinger2012large} and apply it in our BoW codebook generation. KISSME is a bayesian method and only assumes \((\textbf{f}_{i1,j,k1}- \textbf{f}_{i2,j,k2})\) is gaussian distribution, which is quite reasonable in our case. The computation is simple yet the algorithm is effective:
\[\Delta_{\textbf{P}}=\sum_{(\textbf{f}_{i1,j,k1}, \textbf{f}_{i2,j,k2}) \in \textbf{P}} (\textbf{f}_{i1,j,k1}- \textbf{f}_{i2,j,k2}) \cdot (\textbf{f}_{i1,j,k1}- \textbf{f}_{i2,j,k2})^{T}\]
\[\Delta_{\textbf{N}}=\sum_{(\textbf{f}_{i1,j,k1}, \textbf{f}_{i2,j,k2}) \in \textbf{N}} (\textbf{f}_{i1,j,k1}- \textbf{f}_{i2,j,k2}) \cdot (\textbf{f}_{i1,j,k1}- \textbf{f}_{i2,j,k2})^{T}\]
\[\textbf{M}'=\Delta_{\textbf{P}}^{-1}-\Delta_{\textbf{N}}^{-1}.\]

Our codebook can be generated by clustering low-level features under the learned distance metric as above. We collect all the features with background removed. Then k-means clustering is applied based on the optimized Mahalanobis distance metric \(\textbf{M}'\). Finally, we build our codebook on the clustering centers.

Applying our codebook in test phase is straightforward. We first extract low-level features from a novel test image. Then the feature is compared with visual words in the codebook by the trained Mahalanobis distance \(\textbf{M}'\). Finally, the visual word histogram of a pedestrian image strip is calculated and the image descriptor is the concatenation of all stripes in one image.

The image descriptor generated above can be compared directly under Euclidean distance or conventional part two metric learning methods. These part two metric learning methods operate on image descriptor level, while our proposed method operates on low level visual features in part one. We will demonstrate in section \ref{s:experiments} that our proposed method can be directly integrated with these conventional methods with a significant performance boost.

\section{Experiments}
\label{s:experiments}

To evaluate the effectiveness of our method, we conducted experiments on 3 public benchmark datasets: the VIPeR \cite{gray2007evaluating}, the PRID450S \cite{roth14a}, and the Market1501 \cite{zheng2015scalable,zheng2016mars} datasets. The conventional evaluation protocol split the dataset into training and test part. For unsupervised methods evaluation, only test samples are used. The BoW codebook size is set to 350 for each feature. An average of 500 superpixels per image are generated by SLIC method and its compactness parameter is set to 20. Considering re-identification as a ranking problem, the performance is measured in Cumulative Matching Characteristics (CMC).

\subsection{Datasets}

\subsubsection{VIPeR}

The 1264 images which are normalized to 128×48 pixels in the VIPeR dataset are captured from 2 different cameras in outdoor environment, including 632 individuals and 2 images for each person. It is the large variances in viewpoint, pose, resolution, and illumination that makes VIPeR very challenging. In conventional evaluations, the dataset is randomly divided into 2 equal parts, one for training, and the other for testing. In one trial, images are taken as probe sequentially and matched against the opposite camera. 10 trials are repeated and the average result is calculated.

\subsubsection{PRID450S}

450 single-shot image pairs depicting walking humans are captured from 2 disjoint surveillance cameras. Pedestrian bounding boxes are manually labeled with a vertical resolution of 100-150 pixels, while the resolution of original images is 720*576 pixels. Moreover, part-level segmentation is provided describing the following regions: head, torso, legs, carried object at torso level (if any) and carried object below torso (if any). Like VIPeR, we randomly partition the dataset into two equal parts, one for training, and the other for testing. 10 trials are repeated.

\subsubsection{Market1501}

Market1501 consists of 32668 detected person bounding boxes of 1501 individuals captured by 6 cameras (5 high-resolution and 1 low-resolution) with overlaps. Each identity is captured by 2 cameras at least, and may have multiple images in one camera. For each identity in test, one query image in each camera is selected, therefore multiple queries are used for each identity. Note that, the selected 3368 queries are hand-drawn, instead of DPM-detected as in the gallery. The provided fixed training and test set are used under both single-query and multi-query evaluation settings.

\subsection{Exploration of metric learning in BoW codebook generation}

\begin{figure}[!t]
\centering
\includegraphics[width=3.5in]{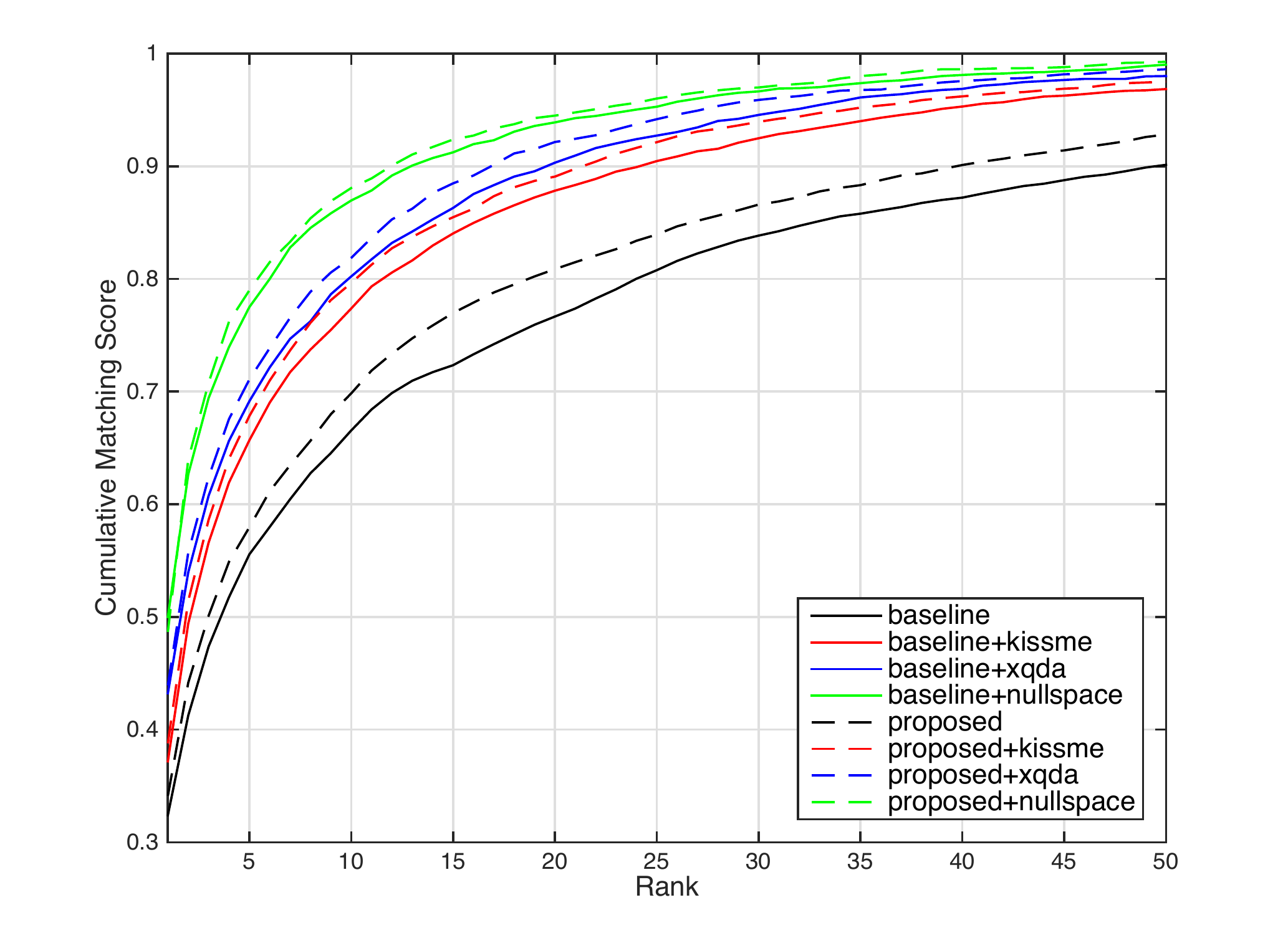}
\caption{CMC curves on the VIPeR dataset, by comparing the proposed approach to conventional baseline methods. Euclidean distance, KISSME, XQDA, and Null Space are employed on image descriptor level respectively.}
\label{f:bow_metric_learning}
\end{figure}

We first compare the performance of our proposed method against conventional baseline BoW approaches on VIPeR dataset. The performance is evaluated on 3 different part two metric learning methods (KISSME \cite{kostinger2012large}, XQDA \cite{liao2015person}, Null Space \cite{zhang2016learning}) on image descriptor level respectively as well as directly applying Euclidean distance on image descriptors without part two metric learning methods. The baseline method applies BoW descriptor simply on Euclidean space without any pedestrian labels, which is totally unsupervised. As shown in Figure \ref{f:bow_metric_learning}, our proposed method performs better than baseline method with 1.7\% rank 1 recognition rate gain. When part two metric learning methods are integrated, the performance gain on rank 1 recognition rate reaches 1.8\% with KISSME metric learning, 0.7\% with XQDA metric learning, and 1.3\% with Null Space metric learning.

The improvement of our proposed method against baseline BoW method is most notable, because the baseline method is totally unsupervised, while the proposed method applies supervised label data on BoW low level feature level. The baseline method with KISSME metric learning outperforms our proposed method without any part two metric learning methods, which suggests that our proposed local feature level metric learning method is an improvement but not replacement of conventional image descriptor level metric learning methods.

\subsection{Comparison to the State-of-the-art results}

In this section, we compare our proposed method with the state-of-the-art approaches. Specifically, we adopt Null Space as the part two image descriptor level metric learning method.

We first compare our approach with the state-of-the-art results on VIPeR in Table \ref{t:viper}. We obtain a rank 1 re-identification rate of 50.0\% on VIPeR, which is superior to the best result by 2.2\%.

\begin{table*}[!t]
\renewcommand{\arraystretch}{1.3}
\caption{Comparison to the State-of-the-Art Results on VIPeR}
\label{t:viper}
\centering
\begin{tabular}{|c||c|c|c|c|c|}
\hline
method & rank 1 & rank 5 & rank 10 & rank 20 & rank 30\\
\hline
SCSP \cite{chen2016similarity} & 53.5 & 82.6 & 91.5 & 96.6 & -\\
\hline
Kernel X-CRC \cite{prates2016kernel} & 51.6 & 80.8 & 89.4 & 95.3 & 97.4\\
\hline
FFN \cite{wu2016enhanced} & 51.1 & 81.0 & 91.4 & 96.9 & -\\
\hline
Triplet Loss \cite{cheng2016person} & 47.8 & 74.7 & 84.8 & 91.1 & 94.3\\
\hline
LSSL \cite{yang2016large} & 47.8 & 77.9 & 87.6 & 94.2 & -\\
\hline
Metric Ensembles \cite{paisitkriangkrai2015learning} & 44.9 & 76.3 & 88.2 & 94.9 & -\\
\hline
LSSCDL \cite{zhang2016sample} & 42.7 & - & 84.3 & 91.9 & -\\
\hline
LOMO + Null Space \cite{zhang2016learning} & 42.3 & 71.5 & 82.9 & 92.1 & -\\
\hline
NLML \cite{huang2015nonlinear} & 42.3 & 71.0 & 85.2 & 94.2 & -\\
\hline
Semantic Representation \cite{shi2015transferring} & 41.6 & 71.9 & 86.2 & 95.1 & -\\
\hline
WARCA \cite{jose2016scalable} & 40.2 & 68.2 & 80.7 & 91.1 & -\\
\hline
LOMO + XQDA \cite{liao2015person} & 40.0 & 68.0 & 80.5 & 91.1 & 95.5\\
\hline
Deep Ranking \cite{chen2016deep} & 38.4 & 69.2 & 81.3 & 90.4 & 94.1\\
\hline
SCNCD \cite{yang2014salient} & 37.8 & 68.5 & 81.2 & 90.4 & 94.2\\
\hline
Correspondence Structure Learning \cite{shen2015person} & 34.8 & 68.7 & 82.3 & 91.8 & 94.9\\
\hline
\bf{Proposed + Null Space} & 50.0 & 79.0 & 88.1 & 94.5 & 97.0\\
\hline
\end{tabular}
\end{table*}

Table \ref{t:prid} compares our results to the state-of-the-art approaches on PRID450S. We yields rank 1 re-identification rate of 70.7\% with Null Space metric learning, which is superior to the best result \cite{prates2016kernel} by 1.9\%.

\begin{table*}[!t]
\renewcommand{\arraystretch}{1.3}
\caption{Comparison to the State-of-the-Art Results on PRID450S}
\label{t:prid}
\centering
\begin{tabular}{|c||c|c|c|c|c|}
\hline
method & rank 1 & rank 5 & rank 10 & rank 20 & rank 30\\
\hline
Kernel X-CRC \cite{prates2016kernel} & 68.8 & 91.2 & 95.9 & 98.4 & 99.0\\
\hline
FFN \cite{wu2016enhanced} & 66.6 & 86.8 & 92.8 & 96.9 & -\\
\hline
LSSCDL \cite{zhang2016sample} & 60.5 & - & 88.6 & 93.6 & -\\
\hline
Semantic Representation \cite{shi2015transferring} & 44.9 & 71.7 & 77.5 & 86.7 & -\\
\hline
Correspondence Structure Learning \cite{shen2015person} & 44.4 & 71.6 & 82.2 & 89.8 & 93.3\\
\hline
SCNCD \cite{yang2014salient} & 41.6 & 68.9 & 79.4 & 87.8 & 95.4\\
\hline
\bf{Proposed + Null Space} & \bf{70.7} & \bf{90.7} & \bf{94.8} & \bf{97.8} & \bf{99.2}\\
\hline
\end{tabular}
\end{table*}

As for the large scale datasets like Market1501, we roughly classify supervised learning methods into two categories, the first conventional metric learning based approaches, and the second deep learning based approaches. Our method yields rank 1 recognition of 64.13\% and mAP of 36.21\% under the single query mode with Null Space \cite{zhang2016learning} metric learning, which outperforms the best metric learning approaches by 8.7\% on rank 1 and 6.3\% on mAP, as shown in Table \ref{tab:market1501}. Our result even outperforms many other deep learning based approaches and is comparable to the recent state-of-the-art method Gated Siamese CNN \cite{varior2016gated}, which is quite outstanding because Market1501 is generally considered more suitable for deep learning based methods with its large image volume.

\begin{table*}[!t]
\renewcommand{\arraystretch}{1.3}
\caption{Comparison to the State-of-the-Art Results on Market1501}
\label{tab:market1501}
\centering
\begin{tabular}{|c|c||c|c|}
\hline
\% & methods & rank 1 & mAP \\
\hline
\multirow{5}{*}{Metric learning}
& WARCA \cite{jose2016scalable} & 45.16 & - \\
& TMA \cite{martinel2016temporal} & 47.92 & 22.31 \\
& SCSP \cite{chen2016similarity} & 51.90 & 26.35 \\
&LOMO+Null Space \cite{zhang2016learning}& 55.43 & 29.87 \\
&\bf{Proposed+Null Space}& \bf{64.13} & \bf{36.21} \\
\hline
\multirow{6}{*}{Deep-learning}
& PersonNet \cite{wu2016personnet} & 37.21 & 18.57 \\
& CAN \cite{liu2016end} & 48.24 & 24.43 \\
& SSDAL \cite{su2016deep} & 39.4 & 19.6 \\
& Triplet CNN \cite{liu2016multi} & 45.1 & - \\
& Histogram Loss \cite{ustinova2016learning} & 59.47 & - \\
& Gated Siamese CNN \cite{varior2016gated} & \bf{65.88} & \bf{39.55} \\
\hline
\end{tabular}
\end{table*}

\section{Conclusion}
\label{s:conclusion}

In this paper, we propose an improved BoW method that learns a suitable metric distance of low level features in codebook generation for person re-identification. The approach uses KISSME metric learning for local features, and can be effectively integrated with conventional image descriptor level metric learning algorithms.  Experiments demonstrate the effectiveness and robustness of our method. The proposed method outperforms state-of-the-art results on VIPeR, PRID450S, and Market1501 integrated with part two Null Space metric learning method.


%


\section*{Acknowledgment}

The work was supported by the National Natural Science Foundation of China under Grant Nos. 61071135 and the National Science and Technology Support Program under Grant No. 2013BAK02B04.

\ifCLASSOPTIONcaptionsoff
  \newpage
\fi


\bibliographystyle{IEEEtranN}
\bibliography{sample}

\begin{thebibliography}{59}
\providecommand{\natexlab}[1]{#1}
\providecommand{\url}[1]{#1}
\csname url@samestyle\endcsname
\providecommand{\newblock}{\relax}
\providecommand{\bibinfo}[2]{#2}
\providecommand{\BIBentrySTDinterwordspacing}{\spaceskip=0pt\relax}
\providecommand{\BIBentryALTinterwordstretchfactor}{4}
\providecommand{\BIBentryALTinterwordspacing}{\spaceskip=\fontdimen2\font plus
\BIBentryALTinterwordstretchfactor\fontdimen3\font minus
  \fontdimen4\font\relax}
\providecommand{\BIBforeignlanguage}[2]{{%
\expandafter\ifx\csname l@#1\endcsname\relax
\typeout{** WARNING: IEEEtranN.bst: No hyphenation pattern has been}%
\typeout{** loaded for the language `#1'. Using the pattern for}%
\typeout{** the default language instead.}%
\else
\language=\csname l@#1\endcsname
\fi
#2}}
\providecommand{\BIBdecl}{\relax}
\BIBdecl

\bibitem[Gong et~al.(2014)Gong, Cristani, Yan, and Loy]{gong2014person}
S.~Gong, M.~Cristani, S.~Yan, and C.~C. Loy, \emph{Person
  re-identification}.\hskip 1em plus 0.5em minus 0.4em\relax Springer, 2014,
  vol.~1.

\bibitem[Lu and Shengjin(2015)]{lu2015person}
T.~Lu and W.~Shengjin, ``Person re-identification as image retrieval using bag
  of ensemble colors,'' \emph{IEICE TRANSACTIONS on Information and Systems},
  vol.~98, no.~1, pp. 180--188, 2015.

\bibitem[K{\"o}stinger et~al.(2012)K{\"o}stinger, Hirzer, Wohlhart, Roth, and
  Bischof]{kostinger2012large}
M.~K{\"o}stinger, M.~Hirzer, P.~Wohlhart, P.~M. Roth, and H.~Bischof, ``Large
  scale metric learning from equivalence constraints,'' in \emph{Computer
  Vision and Pattern Recognition (CVPR), 2012 IEEE Conference on}.\hskip 1em
  plus 0.5em minus 0.4em\relax IEEE, 2012, pp. 2288--2295.

\bibitem[Zhang et~al.(2016{\natexlab{a}})Zhang, Xiang, and
  Gong]{zhang2016learning}
L.~Zhang, T.~Xiang, and S.~Gong, ``Learning a discriminative null space for
  person re-identification,'' \emph{arXiv preprint arXiv:1603.02139}, 2016.

\bibitem[Zheng et~al.(2016{\natexlab{a}})Zheng, Yang, and
  Hauptmann]{zheng2016person}
L.~Zheng, Y.~Yang, and A.~G. Hauptmann, ``Person re-identification: Past,
  present and future,'' \emph{arXiv preprint arXiv:1610.02984}, 2016.

\bibitem[Gray and Tao(2008)]{gray2008viewpoint}
D.~Gray and H.~Tao, ``Viewpoint invariant pedestrian recognition with an
  ensemble of localized features,'' in \emph{European conference on computer
  vision}.\hskip 1em plus 0.5em minus 0.4em\relax Springer, 2008, pp. 262--275.

\bibitem[Farenzena et~al.(2010)Farenzena, Bazzani, Perina, Murino, and
  Cristani]{farenzena2010person}
M.~Farenzena, L.~Bazzani, A.~Perina, V.~Murino, and M.~Cristani, ``Person
  re-identification by symmetry-driven accumulation of local features,'' in
  \emph{Computer Vision and Pattern Recognition (CVPR), 2010 IEEE Conference
  on}.\hskip 1em plus 0.5em minus 0.4em\relax IEEE, 2010, pp. 2360--2367.

\bibitem[Zhao et~al.(2013)Zhao, Ouyang, and Wang]{zhao2013unsupervised}
R.~Zhao, W.~Ouyang, and X.~Wang, ``Unsupervised salience learning for person
  re-identification,'' in \emph{Proceedings of the IEEE Conference on Computer
  Vision and Pattern Recognition}, 2013, pp. 3586--3593.

\bibitem[Das et~al.(2014)Das, Chakraborty, and
  Roy-Chowdhury]{das2014consistent}
A.~Das, A.~Chakraborty, and A.~K. Roy-Chowdhury, ``Consistent re-identification
  in a camera network,'' in \emph{European Conference on Computer
  Vision}.\hskip 1em plus 0.5em minus 0.4em\relax Springer, 2014, pp. 330--345.

\bibitem[Li et~al.(2013)Li, Chang, Liang, Huang, Cao, and
  Smith]{li2013learning}
Z.~Li, S.~Chang, F.~Liang, T.~S. Huang, L.~Cao, and J.~R. Smith, ``Learning
  locally-adaptive decision functions for person verification,'' in
  \emph{Proceedings of the IEEE Conference on Computer Vision and Pattern
  Recognition}, 2013, pp. 3610--3617.

\bibitem[Zhou et~al.(2009)Zhou, Cui, Li, Liang, and
  Huang]{zhou2009hierarchical}
X.~Zhou, N.~Cui, Z.~Li, F.~Liang, and T.~S. Huang, ``Hierarchical
  gaussianization for image classification,'' in \emph{2009 IEEE 12th
  International Conference on Computer Vision}.\hskip 1em plus 0.5em minus
  0.4em\relax IEEE, 2009, pp. 1971--1977.

\bibitem[Chen et~al.(2015)Chen, Yuan, Hua, Zheng, and Wang]{chen2015similarity}
D.~Chen, Z.~Yuan, G.~Hua, N.~Zheng, and J.~Wang, ``Similarity learning on an
  explicit polynomial kernel feature map for person re-identification,'' in
  \emph{Proceedings of the IEEE Conference on Computer Vision and Pattern
  Recognition}, 2015, pp. 1565--1573.

\bibitem[Pedagadi et~al.(2013)Pedagadi, Orwell, Velastin, and
  Boghossian]{pedagadi2013local}
S.~Pedagadi, J.~Orwell, S.~Velastin, and B.~Boghossian, ``Local fisher
  discriminant analysis for pedestrian re-identification,'' in
  \emph{Proceedings of the IEEE Conference on Computer Vision and Pattern
  Recognition}, 2013, pp. 3318--3325.

\bibitem[Liu et~al.(2014)Liu, Song, Tao, Zhou, Chen, and Bu]{liu2014semi}
X.~Liu, M.~Song, D.~Tao, X.~Zhou, C.~Chen, and J.~Bu, ``Semi-supervised coupled
  dictionary learning for person re-identification,'' in \emph{Proceedings of
  the IEEE Conference on Computer Vision and Pattern Recognition}, 2014, pp.
  3550--3557.

\bibitem[Yang et~al.(2014)Yang, Yang, Yan, Liao, Yi, and Li]{yang2014salient}
Y.~Yang, J.~Yang, J.~Yan, S.~Liao, D.~Yi, and S.~Z. Li, ``Salient color names
  for person re-identification,'' in \emph{European Conference on Computer
  Vision}.\hskip 1em plus 0.5em minus 0.4em\relax Springer, 2014, pp. 536--551.

\bibitem[Liao et~al.(2015)Liao, Hu, Zhu, and Li]{liao2015person}
S.~Liao, Y.~Hu, X.~Zhu, and S.~Z. Li, ``Person re-identification by local
  maximal occurrence representation and metric learning,'' in \emph{Proceedings
  of the IEEE Conference on Computer Vision and Pattern Recognition}, 2015, pp.
  2197--2206.

\bibitem[Van~de Weijer et~al.(2007)Van~de Weijer, Schmid, and
  Verbeek]{van2007learning}
J.~Van~de Weijer, C.~Schmid, and J.~Verbeek, ``Learning color names from
  real-world images,'' in \emph{2007 IEEE Conference on Computer Vision and
  Pattern Recognition}.\hskip 1em plus 0.5em minus 0.4em\relax IEEE, 2007, pp.
  1--8.

\bibitem[Weinberger and Saul(2009)]{weinberger2009distance}
K.~Q. Weinberger and L.~K. Saul, ``Distance metric learning for large margin
  nearest neighbor classification,'' \emph{Journal of Machine Learning
  Research}, vol.~10, no. Feb, pp. 207--244, 2009.

\bibitem[Hirzer et~al.(2012)Hirzer, Roth, K{\"o}stinger, and
  Bischof]{hirzer2012relaxed}
M.~Hirzer, P.~M. Roth, M.~K{\"o}stinger, and H.~Bischof, ``Relaxed pairwise
  learned metric for person re-identification,'' in \emph{European Conference
  on Computer Vision}.\hskip 1em plus 0.5em minus 0.4em\relax Springer, 2012,
  pp. 780--793.

\bibitem[Liao and Li(2015)]{liao2015efficient}
S.~Liao and S.~Z. Li, ``Efficient psd constrained asymmetric metric learning
  for person re-identification,'' in \emph{Proceedings of the IEEE
  International Conference on Computer Vision}, 2015, pp. 3685--3693.

\bibitem[Scholkopft and Mullert(1999)]{scholkopft1999fisher}
B.~Scholkopft and K.-R. Mullert, ``Fisher discriminant analysis with kernels,''
  \emph{Neural networks for signal processing IX}, vol.~1, no.~1, p.~1, 1999.

\bibitem[Li et~al.(2014)Li, Zhao, Xiao, and Wang]{li2014deepreid}
W.~Li, R.~Zhao, T.~Xiao, and X.~Wang, ``Deepreid: Deep filter pairing neural
  network for person re-identification,'' in \emph{Proceedings of the IEEE
  Conference on Computer Vision and Pattern Recognition}, 2014, pp. 152--159.

\bibitem[Ahmed et~al.(2015)Ahmed, Jones, and Marks]{ahmed2015improved}
E.~Ahmed, M.~Jones, and T.~K. Marks, ``An improved deep learning architecture
  for person re-identification,'' in \emph{Proceedings of the IEEE Conference
  on Computer Vision and Pattern Recognition}, 2015, pp. 3908--3916.

\bibitem[Yi et~al.(2014)Yi, Lei, Liao, and Li]{yi2014deep}
D.~Yi, Z.~Lei, S.~Liao, and S.~Z. Li, ``Deep metric learning for person
  re-identification,'' in \emph{Pattern Recognition (ICPR), 2014 22nd
  International Conference on}.\hskip 1em plus 0.5em minus 0.4em\relax IEEE,
  2014, pp. 34--39.

\bibitem[Ding et~al.(2015)Ding, Lin, Wang, and Chao]{ding2015deep}
S.~Ding, L.~Lin, G.~Wang, and H.~Chao, ``Deep feature learning with relative
  distance comparison for person re-identification,'' \emph{Pattern
  Recognition}, vol.~48, no.~10, pp. 2993--3003, 2015.

\bibitem[Zheng et~al.(2016{\natexlab{b}})Zheng, Zheng, and
  Yang]{zheng2016discriminatively}
Z.~Zheng, L.~Zheng, and Y.~Yang, ``A discriminatively learned cnn embedding for
  person re-identification,'' \emph{arXiv preprint arXiv:1611.05666}, 2016.

\bibitem[Zheng et~al.(2017)Zheng, Huang, Lu, and Yang]{zheng2017pose}
L.~Zheng, Y.~Huang, H.~Lu, and Y.~Yang, ``Pose invariant embedding for deep
  person re-identification,'' \emph{arXiv preprint arXiv:1701.07732}, 2017.

\bibitem[Zheng et~al.(2015)Zheng, Shen, Tian, Wang, Wang, and
  Tian]{zheng2015scalable}
L.~Zheng, L.~Shen, L.~Tian, S.~Wang, J.~Wang, and Q.~Tian, ``Scalable person
  re-identification: A benchmark,'' in \emph{Computer Vision, IEEE
  International Conference on}, 2015.

\bibitem[Achanta et~al.(2012)Achanta, Shaji, Smith, Lucchi, Fua, and
  S{\"u}sstrunk]{achanta2012slic}
R.~Achanta, A.~Shaji, K.~Smith, A.~Lucchi, P.~Fua, and S.~S{\"u}sstrunk, ``Slic
  superpixels compared to state-of-the-art superpixel methods,'' \emph{IEEE
  transactions on pattern analysis and machine intelligence}, vol.~34, no.~11,
  pp. 2274--2282, 2012.

\bibitem[Gray(1984)]{gray1984vector}
R.~Gray, ``Vector quantization,'' \emph{IEEE Assp Magazine}, vol.~1, no.~2, pp.
  4--29, 1984.

\bibitem[Jegou et~al.(2008)Jegou, Douze, and Schmid]{jegou2008hamming}
H.~Jegou, M.~Douze, and C.~Schmid, ``Hamming embedding and weak geometric
  consistency for large scale image search,'' in \emph{European conference on
  computer vision}.\hskip 1em plus 0.5em minus 0.4em\relax Springer, 2008, pp.
  304--317.

\bibitem[Luo et~al.(2013)Luo, Wang, and Tang]{luo2013pedestrian}
P.~Luo, X.~Wang, and X.~Tang, ``Pedestrian parsing via deep decompositional
  network,'' in \emph{Proceedings of the IEEE International Conference on
  Computer Vision}, 2013, pp. 2648--2655.

\bibitem[Berlin and Kay(1991)]{berlin1991basic}
B.~Berlin and P.~Kay, \emph{Basic color terms: Their universality and
  evolution}.\hskip 1em plus 0.5em minus 0.4em\relax Univ of California Press,
  1991.

\bibitem[Dalal and Triggs(2005)]{dalal2005histograms}
N.~Dalal and B.~Triggs, ``Histograms of oriented gradients for human
  detection,'' in \emph{2005 IEEE Computer Society Conference on Computer
  Vision and Pattern Recognition (CVPR'05)}, vol.~1.\hskip 1em plus 0.5em minus
  0.4em\relax IEEE, 2005, pp. 886--893.

\bibitem[Liao et~al.(2010)Liao, Zhao, Kellokumpu, Pietik{\"a}inen, and
  Li]{liao2010modeling}
S.~Liao, G.~Zhao, V.~Kellokumpu, M.~Pietik{\"a}inen, and S.~Z. Li, ``Modeling
  pixel process with scale invariant local patterns for background subtraction
  in complex scenes,'' in \emph{Computer Vision and Pattern Recognition (CVPR),
  2010 IEEE Conference on}.\hskip 1em plus 0.5em minus 0.4em\relax IEEE, 2010,
  pp. 1301--1306.

\bibitem[Arandjelovi{\'c} and Zisserman(2012)]{arandjelovic2012three}
R.~Arandjelovi{\'c} and A.~Zisserman, ``Three things everyone should know to
  improve object retrieval,'' in \emph{Computer Vision and Pattern Recognition
  (CVPR), 2012 IEEE Conference on}.\hskip 1em plus 0.5em minus 0.4em\relax
  IEEE, 2012, pp. 2911--2918.

\bibitem[Ojala et~al.(1996)Ojala, Pietik{\"a}inen, and
  Harwood]{ojala1996comparative}
T.~Ojala, M.~Pietik{\"a}inen, and D.~Harwood, ``A comparative study of texture
  measures with classification based on featured distributions,'' \emph{Pattern
  recognition}, vol.~29, no.~1, pp. 51--59, 1996.

\bibitem[Gray et~al.(2007)Gray, Brennan, and Tao]{gray2007evaluating}
D.~Gray, S.~Brennan, and H.~Tao, ``Evaluating appearance models for
  recognition, reacquisition, and tracking,'' in \emph{Proc. IEEE International
  Workshop on Performance Evaluation for Tracking and Surveillance (PETS)},
  vol.~3, no.~5.\hskip 1em plus 0.5em minus 0.4em\relax Citeseer, 2007.

\bibitem[Roth et~al.(2014)Roth, Hirzer, Koestinger, Beleznai, and
  Bischof]{roth14a}
P.~M. Roth, M.~Hirzer, M.~Koestinger, C.~Beleznai, and H.~Bischof,
  ``Mahalanobis distance learning for person re-identification,'' in
  \emph{Person Re-Identification}, ser. Advances in Computer Vision and Pattern
  Recognition, S.~Gong, M.~Cristani, S.~Yan, and C.~C. Loy, Eds.\hskip 1em plus
  0.5em minus 0.4em\relax London, United Kingdom: Springer, 2014, pp. 247--267.

\bibitem[Zheng et~al.(2016{\natexlab{c}})Zheng, Bie, Sun, Wang, Su, Wang, and
  Tian]{zheng2016mars}
L.~Zheng, Z.~Bie, Y.~Sun, J.~Wang, C.~Su, S.~Wang, and Q.~Tian, ``Mars: A video
  benchmark for large-scale person re-identification,'' in \emph{European
  Conference on Computer Vision}.\hskip 1em plus 0.5em minus 0.4em\relax
  Springer, 2016, pp. 868--884.

\bibitem[Chen et~al.(2016{\natexlab{a}})Chen, Yuan, Chen, and
  Zheng]{chen2016similarity}
D.~Chen, Z.~Yuan, B.~Chen, and N.~Zheng, ``Similarity learning with spatial
  constraints for person re-identification,'' in \emph{Proceedings of the IEEE
  Conference on Computer Vision and Pattern Recognition}, 2016, pp. 1268--1277.

\bibitem[Prates and Schwartz(2016)]{prates2016kernel}
R.~Prates and W.~R. Schwartz, ``Kernel cross-view collaborative representation
  based classification for person re-identification,'' \emph{arXiv preprint
  arXiv:1611.06969}, 2016.

\bibitem[Wu et~al.(2016{\natexlab{a}})Wu, Chen, Li, Wu, You, and
  Zheng]{wu2016enhanced}
S.~Wu, Y.-C. Chen, X.~Li, A.-C. Wu, J.-J. You, and W.-S. Zheng, ``An enhanced
  deep feature representation for person re-identification,'' in \emph{2016
  IEEE Winter Conference on Applications of Computer Vision (WACV)}.\hskip 1em
  plus 0.5em minus 0.4em\relax IEEE, 2016, pp. 1--8.

\bibitem[Cheng et~al.(2016)Cheng, Gong, Zhou, Wang, and Zheng]{cheng2016person}
D.~Cheng, Y.~Gong, S.~Zhou, J.~Wang, and N.~Zheng, ``Person re-identification
  by multi-channel parts-based cnn with improved triplet loss function,'' in
  \emph{Proceedings of the IEEE Conference on Computer Vision and Pattern
  Recognition}, 2016, pp. 1335--1344.

\bibitem[Yang et~al.(2016)Yang, Liao, Lei, and Li]{yang2016large}
Y.~Yang, S.~Liao, Z.~Lei, and S.~Z. Li, ``Large scale similarity learning using
  similar pairs for person verification,'' in \emph{Thirtieth AAAI Conference
  on Artificial Intelligence}, 2016.

\bibitem[Paisitkriangkrai et~al.(2015)Paisitkriangkrai, Shen, and van~den
  Hengel]{paisitkriangkrai2015learning}
S.~Paisitkriangkrai, C.~Shen, and A.~van~den Hengel, ``Learning to rank in
  person re-identification with metric ensembles,'' in \emph{Proceedings of the
  IEEE Conference on Computer Vision and Pattern Recognition}, 2015, pp.
  1846--1855.

\bibitem[Zhang et~al.(2016{\natexlab{b}})Zhang, Li, Lu, Irie, and
  Ruan]{zhang2016sample}
Y.~Zhang, B.~Li, H.~Lu, A.~Irie, and X.~Ruan, ``Sample-specific svm learning
  for person re-identification,'' in \emph{Proceedings of the IEEE Conference
  on Computer Vision and Pattern Recognition}, 2016.

\bibitem[Huang et~al.(2015)Huang, Lu, Zhou, and Jain]{huang2015nonlinear}
S.~Huang, J.~Lu, J.~Zhou, and A.~K. Jain, ``Nonlinear local metric learning for
  person re-identification,'' \emph{arXiv preprint arXiv:1511.05169}, 2015.

\bibitem[Shi et~al.(2015)Shi, Hospedales, and Xiang]{shi2015transferring}
Z.~Shi, T.~M. Hospedales, and T.~Xiang, ``Transferring a semantic
  representation for person re-identification and search,'' in
  \emph{Proceedings of the IEEE Conference on Computer Vision and Pattern
  Recognition}, 2015, pp. 4184--4193.

\bibitem[Jose and Fleuret(2016)]{jose2016scalable}
C.~Jose and F.~Fleuret, ``Scalable metric learning via weighted approximate
  rank component analysis,'' \emph{arXiv preprint arXiv:1603.00370}, 2016.

\bibitem[Chen et~al.(2016{\natexlab{b}})Chen, Guo, and Lai]{chen2016deep}
S.-Z. Chen, C.-C. Guo, and J.-H. Lai, ``Deep ranking for person
  re-identification via joint representation learning,'' \emph{IEEE
  Transactions on Image Processing}, vol.~25, no.~5, pp. 2353--2367, 2016.

\bibitem[Shen et~al.(2015)Shen, Lin, Yan, Xu, Wu, and Wang]{shen2015person}
Y.~Shen, W.~Lin, J.~Yan, M.~Xu, J.~Wu, and J.~Wang, ``Person re-identification
  with correspondence structure learning,'' in \emph{Proceedings of the IEEE
  International Conference on Computer Vision}, 2015, pp. 3200--3208.

\bibitem[Varior et~al.(2016)Varior, Haloi, and Wang]{varior2016gated}
R.~R. Varior, M.~Haloi, and G.~Wang, ``Gated siamese convolutional neural
  network architecture for human re-identification,'' in \emph{European
  Conference on Computer Vision}.\hskip 1em plus 0.5em minus 0.4em\relax
  Springer, 2016, pp. 791--808.

\bibitem[Martinel et~al.(2016)Martinel, Das, Micheloni, and
  Roy-Chowdhury]{martinel2016temporal}
N.~Martinel, A.~Das, C.~Micheloni, and A.~K. Roy-Chowdhury, ``Temporal model
  adaptation for person re-identification,'' in \emph{European Conference on
  Computer Vision}.\hskip 1em plus 0.5em minus 0.4em\relax Springer, 2016, pp.
  858--877.

\bibitem[Wu et~al.(2016{\natexlab{b}})Wu, Shen, and Hengel]{wu2016personnet}
L.~Wu, C.~Shen, and A.~v.~d. Hengel, ``Personnet: Person re-identification with
  deep convolutional neural networks,'' \emph{arXiv preprint arXiv:1601.07255},
  2016.

\bibitem[Liu et~al.(2016{\natexlab{a}})Liu, Feng, Qi, Jiang, and
  Yan]{liu2016end}
H.~Liu, J.~Feng, M.~Qi, J.~Jiang, and S.~Yan, ``End-to-end comparative
  attention networks for person re-identification,'' \emph{arXiv preprint
  arXiv:1606.04404}, 2016.

\bibitem[Su et~al.(2016)Su, Zhang, Xing, Gao, and Tian]{su2016deep}
C.~Su, S.~Zhang, J.~Xing, W.~Gao, and Q.~Tian, ``Deep attributes driven
  multi-camera person re-identification,'' \emph{arXiv preprint
  arXiv:1605.03259}, 2016.

\bibitem[Liu et~al.(2016{\natexlab{b}})Liu, Zha, Tian, Liu, Yao, Ling, and
  Mei]{liu2016multi}
J.~Liu, Z.-J. Zha, Q.~Tian, D.~Liu, T.~Yao, Q.~Ling, and T.~Mei, ``Multi-scale
  triplet cnn for person re-identification,'' in \emph{Proceedings of the 2016
  ACM on Multimedia Conference}.\hskip 1em plus 0.5em minus 0.4em\relax ACM,
  2016, pp. 192--196.

\bibitem[Ustinova and Lempitsky(2016)]{ustinova2016learning}
E.~Ustinova and V.~Lempitsky, ``Learning deep embeddings with histogram loss,''
  in \emph{Advances In Neural Information Processing Systems}, 2016, pp.
  4170--4178.

\end{thebibliography}

%

\begin{IEEEbiographynophoto}{Lu Tian}
was born in 1989. She received the bachelor degree in Electronic Engineering from Tsinghua University, China, in 2011. She has been studying for the Ph.D. degree in Electronic Engineering of Tsinghua University from 2011. Her current research interests include pattern recognition, human feature extraction, in particular person re-identification.
\end{IEEEbiographynophoto}

\begin{IEEEbiographynophoto}{Shengjin Wang}
received the B.E. degree from Tsinghua University, China, and the Ph.D. degree from the Tokyo Institute of Technology, Tokyo, Japan, in 1985 and 1997, respectively. From 1997 to 2003, he was a member of the researcher with the Internet System Research Laboratories, NEC Corporation, Japan. Since 2003, he has been a Professor with the Department of Electronic Engineering, Tsinghua University, where he is currently the Director of the Research Institute of Image and Graphics. His current research interests include image processing, computer vision, video surveillance, and pattern recognition. He is a member of IEEE and IEICE.
\end{IEEEbiographynophoto}




\end{document}